\documentclass[letterpaper, 10 pt, conference]{ieeeconf}  

\IEEEoverridecommandlockouts                              

\usepackage{floatrow}

\usepackage{graphicx}

\usepackage{color}
\usepackage{enumerate}
\usepackage{amsthm}
\usepackage{amsfonts}
\usepackage{amsmath}
\usepackage{amssymb}
\usepackage[noend]{algorithmic}
\usepackage{xcolor}
\usepackage{prettyref}
\usepackage{array,makecell} 
\usepackage{arydshln} 
\usepackage{caption} 
\usepackage{todonotes} 

\usepackage{etoolbox}
\makeatletter
\patchcmd{\@makecaption}
  {\scshape}
  {}
  {}
  {}
\makeatother

\makeatletter
\let\NAT@parse\undefined
\makeatother
\usepackage[numbers,sort&compress,sectionbib]{natbib} 

\usepackage[breaklinks=true]{hyperref}
\usepackage{xspace} 

\newrefformat{sec}{Section~\ref{#1}}
\newrefformat{tab}{Table~\ref{#1}}
\newrefformat{fig}{Figure~\ref{#1}}

\newcommand{\ie}{{\em i.e.}\xspace}
\newcommand{\eg}{{\em e.g.}\xspace}

\newcommand{\etal}{{\em et al.}\xspace}

\newcommand{\abs}[1]{|#1|\xspace}
\newcommand{\inputD}{D}
\newcommand{\gtD}{D^*}
\newcommand{\cart}[1]{\mathbf{card}\left( #1 \right)}

\newcommand{\RGB}{\texttt{RGB}\xspace}
\newcommand{\RGBd}{\texttt{RGBd}\xspace}
\newcommand{\sd}{\texttt{sd}\xspace}

\newcommand{\nyudepth}{NYU-Depth-v2\xspace}
\newcommand{\makethreed}{Make3D\xspace}
\newcommand{\kitti}{KITTI\xspace}

\newcommand{\berhu}{berHu\xspace}
\newcommand{\berhumath}{\mathcal{B}\xspace}
\newcommand{\ltwo}{$\mathcal{L}_2$\xspace}
\newcommand{\lone}{$\mathcal{L}_1$\xspace}

\newcommand{\conv}{$\mathrm{Conv}$\xspace}
\newcommand{\channeldrop}{$\mathrm{ChanDrop}$\xspace}
\newcommand{\depthwise}{$\mathrm{DepthWise}$\xspace}

\newcommand{\deconv}[1]{$\mathrm{DeConv}_{#1}$\xspace}
\newcommand{\upconv}{$\mathrm{UpConv}$\xspace}
\newcommand{\upproj}{$\mathrm{UpProj}$\xspace}

\newcommand{\rmse}{$\mathrm{RMSE}$\xspace}

\newcommand{\absrel}{$\mathrm{REL}$\xspace}
\newcommand{\deltas}[1]{$\delta_{#1}$\xspace}

\title{\LARGE \bf
Sparse-to-Dense: Depth Prediction from\\Sparse Depth Samples and a Single Image
}

\author{Fangchang Ma$^{1}$ and Sertac Karaman$^{1}$
\thanks{$^{1}$F. Ma and S. Karaman are with 
  the Laboratory for Information \& Decision Systems, Massachusetts 
  Institute of Technology, Cambridge, MA, USA.
        {\tt\small \{fcma, sertac\}@mit.edu}}%
\thanks{$^{2}${\url{https://github.com/fangchangma/sparse-to-dense}}}
\thanks{$^{3}${\url{https://www.youtube.com/watch?v=vNIIT_M7x7Y}}}
}

\begin{document}
\bstctlcite{IEEEexample:BSTcontrol} 

\maketitle
\thispagestyle{empty}
\pagestyle{empty}

\begin{abstract}

We consider the problem of dense depth prediction from a sparse set of depth measurements and a single RGB image. Since depth estimation from monocular images alone is inherently ambiguous and unreliable, to attain a higher level of robustness and accuracy, we introduce additional sparse depth samples, which are either acquired with a low-resolution depth sensor or computed via visual Simultaneous Localization and Mapping (SLAM) algorithms.
We propose the use of a single deep regression network to learn directly from the RGB-D raw data, and explore the impact of number of depth samples on prediction accuracy.
Our experiments show that, compared to using only RGB images, the addition of 100 spatially random depth samples reduces the prediction root-mean-square error by 50\% on the \nyudepth indoor dataset. It also boosts the percentage of reliable prediction from 59\% to 92\% on the \kitti dataset.
We demonstrate two applications of the proposed algorithm: a plug-in module in SLAM to convert sparse maps to dense maps, and super-resolution for LiDARs.
Software$^{2}$ and video demonstration$^{3}$ are publicly available.

\end{abstract}


\section{Introduction}

Depth sensing and estimation is of vital importance in a wide range of engineering applications, such as robotics, autonomous driving, augmented reality (AR) and 3D mapping. 
However, existing depth sensors, including LiDARs, structured-light-based depth sensors, and stereo cameras, all have their own limitations.
For instance, the top-of-the-range 3D LiDARs are cost-prohibitive (with up to \$75,000 cost per unit), and yet provide only sparse measurements for distant objects. 
Structured-light-based depth sensors (\eg Kinect) are sunlight-sensitive and power-consuming, with a short ranging distance.
Finally, stereo cameras require a large baseline and careful calibration for accurate triangulation, which demands large amount of computation and usually fails at featureless regions.
Because of these limitations, there has always been a strong interest in depth estimation using a single camera, which is small, low-cost, energy-efficient, and ubiquitous in consumer electronic products. 

However, the accuracy and reliability of such methods is still far from being practical,
despite over a decade of research effort devoted to RGB-based depth prediction including the recent improvements with deep learning approaches. For instance, the state-of-the-art RGB-based depth prediction methods~\cite{liu2015deep,eigen2015predicting,laina2016deeper} produce an average error (measured by the root mean squared error) of over 50cm in indoor scenarios (\eg, on the \nyudepth dataset~\cite{silberman2012indoor}). Such methods perform even worse outdoors, with at least 4 meters of average error on \makethreed and \kitti datasets~\cite{saxena2009make3d,Geiger2012CVPR}. 


\begin{figure}[t]
\centering
\begin{minipage}{\textwidth}
\newcommand{\figWidth}{ 0.485\linewidth } 
\setlength\tabcolsep{0.4mm} 
\begin{tabular}{ c c }
  \begin{minipage}[m]{\figWidth}\centering
  \includegraphics[width=\linewidth]{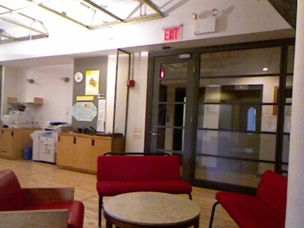} \\
  \vspace{-0.1cm}
  (a) RGB
  \end{minipage}
  & 
  \begin{minipage}[m]{\figWidth}\centering
  \includegraphics[width=\linewidth,height=0.75\linewidth]{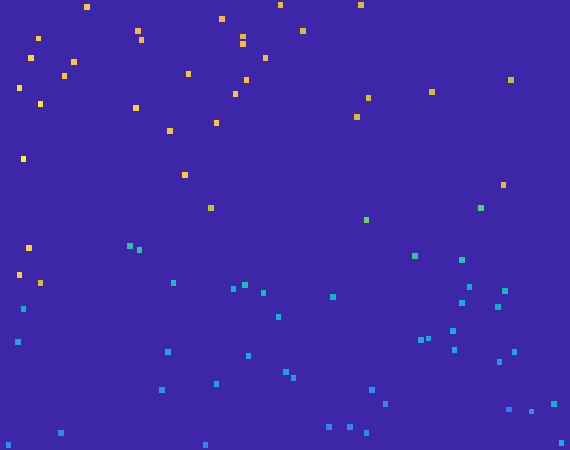} \\
  \vspace{-0.1cm}
  (b) sparse depth
  \end{minipage}
  \\
  \begin{minipage}[m]{\figWidth}\centering
  \includegraphics[width=\linewidth]{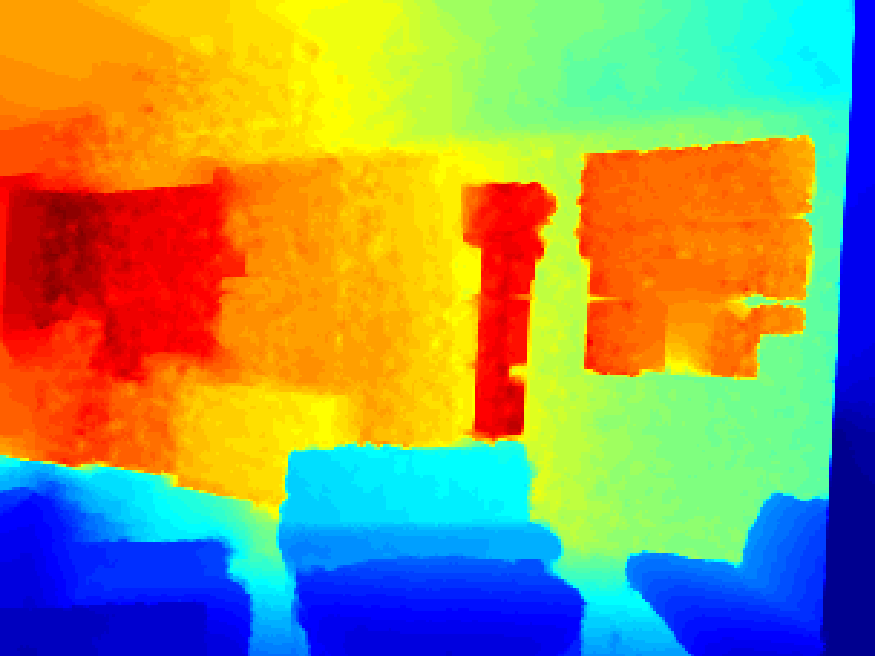} \\
  \vspace{-0.1cm}
  (c) ground truth
  \end{minipage}
  & 
  \begin{minipage}[m]{\figWidth}\centering
  \includegraphics[width=\linewidth]{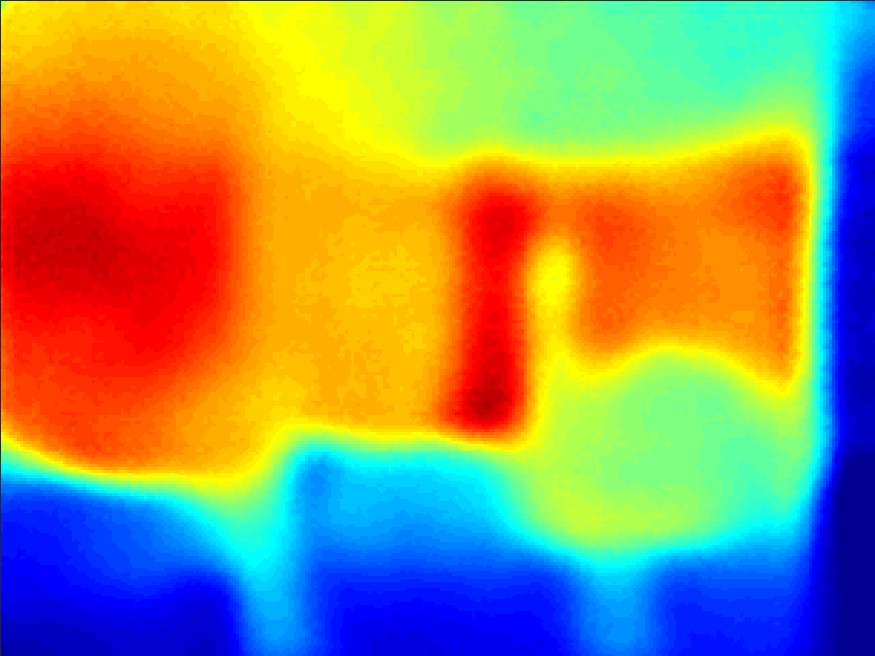} \\
  \vspace{-0.1cm}
  (d) prediction
  \end{minipage}
\end{tabular}
\end{minipage}

\caption{We develop a deep regression model to predict dense depth image from a single RGB image and a set of sparse depth samples. Our method significantly outperforms RGB-based and other fusion-based algorithms.}
\label{fig:RGBd}
\end{figure}

To address the potential fundamental limitations of RGB-based depth estimation, we consider the utilization of sparse depth measurements, along with RGB data, to reconstruct depth in full resolution. Sparse depth measurements are readily available in many applications. For instance, low-resolution depth sensors (\eg, a low-cost LiDARs) provide such measurements. Sparse depth measurements can also be computed from the output of SLAM\footnotemark[4] and visual-inertial odometry algorithms.
\footnotetext[4]{A typical feature-based SLAM algorithm, such as ORB-SLAM~\cite{mur2015orb}, keeps track of hundreds of 3D landmarks in each frame.}
In this work, we demonstrate the effectiveness of using sparse depth measurements, in addition to the RGB images, as part of the input to the system. We use a single convolutional neural network to learn a deep regression model for depth image prediction. Our experimental results show that the addition of as few as 100 depth samples reduces the root mean squared error by over 50\% on the \nyudepth dataset, and boosts the percentage of reliable prediction from 59\% to 92\% on the more challenging \kitti outdoor dataset. 
In general, our results show that the addition of a few sparse depth samples drastically improves depth reconstruction performance. Our quantitative results may help inform the development of sensors for future robotic vehicles and consumer devices. 

The main contribution of this paper is a deep regression model that takes both a sparse set of depth samples and RGB images as input and predicts a full-resolution depth image. The prediction accuracy of our method significantly outperforms state-of-the-art methods, including both RGB-based and fusion-based techniques. Furthermore, we demonstrate in experiments that our method can be used as a plug-in module to sparse visual odometry / SLAM algorithms to create an accurate, dense point cloud. In addition, we show that our method can also be used in 3D LiDARs to create much denser measurements.


\section{Related Work}


{\bf RGB-based depth prediction} Early works on depth estimation using RGB images usually relied on hand-crafted features and probabilistic graphical models. For instance, \citet{saxena2006learning} estimated the absolute scales of different image patches and inferred the depth image using a Markov Random Field model. Non-parametric approaches~\cite{karsch2012depth,konrad20122d,karsch2014depthtransfer,liu2014discrete} were also exploited to estimate the depth of a query image by combining the depths of images with similar photometric content retrieved from a database. 

Recently, deep learning has been successfully applied to the depth estimation problem. \citet{eigen2014depth} suggest a two-stack convolutional neural network (CNN), with one predicting the global coarse scale and the other refining local details. \citet{eigen2015predicting} further incorporate other auxiliary prediction tasks into the same architecture. \citet{liu2015deep} combined a deep CNN and a continuous conditional random field, and attained visually sharper transitions and local details. 
\citet{laina2016deeper} developed a deep residual network based on the ResNet~\cite{he2016deep} and achieved higher accuracy than \cite{liu2015deep,eigen2015predicting}. Semi-supervised~\cite{kuznietsov2017semi} and unsupervised learning~\cite{zhou2017unsupervised,garg2016unsupervised,godard2016unsupervised} setups have also been explored for disparity image prediction. For instance, \citet{godard2016unsupervised} formulated disparity estimation as an image reconstruction problem, where neural networks were trained to warp left images to match the right.



{\bf Depth reconstruction from sparse samples} Another line of related work is depth reconstruction from sparse samples. A common ground of many approaches in this area is the use of sparse representations for depth signals. For instance, \citet{hawe2011dense} assumed that disparity maps were sparse on the Wavelet basis and reconstructed a dense disparity image with a conjugate sub-gradient method. \citet{liu2015depth} combined wavelet and contourlet dictionaries for more accurate reconstruction. Our previous work on sparse depth sensing~\cite{ma2016sparse, ma2017sparse} exploited the sparsity underlying the second-order derivatives of depth images, and outperformed both \cite{hawe2011dense,liu2015deep} in reconstruction accuracy and speed. 

{\bf Sensor fusion}
A wide range of techniques attempted to improve depth prediction by fusing additional information from different sensor modalities. For instance, 
\citet{mancini2016fast} proposed a CNN that took both RGB images and optical flow images as input to predict distance. 
\citet{liao2017parse} studied the use of a 2D laser scanner mounted on a mobile ground robot to provide an additional reference depth signal as input and obtained higher accuracy than using RGB images alone. Compared to the approach by \citet{liao2017parse}, this work makes no assumption regarding the orientation or position of sensors, nor the spatial distribution of input depth samples in the pixel space. 
\citet{cadena2016multi} developed a multi-modal auto-encoder to learn from three input modalities, including RGB, depth, and semantic labels. In their experiments, \citet{cadena2016multi} used sparse depth on extracted FAST corner features as part of the input to the system to produce a low-resolution depth prediction. The accuracy was comparable to using RGB alone. In comparison, our method predicts a full-resolution depth image, learns a better cross-modality representation for RGB and sparse depth, and attains a significantly higher accuracy.




\section{Methodology}

In this section, we describe the architecture of the convolutional neural network. We also discuss the depth sampling strategy, the data augmentation techniques, and the loss functions used for training.

\subsection{CNN Architecture}
\begin{figure*}[h]
\centering
\includegraphics[width=\textwidth]{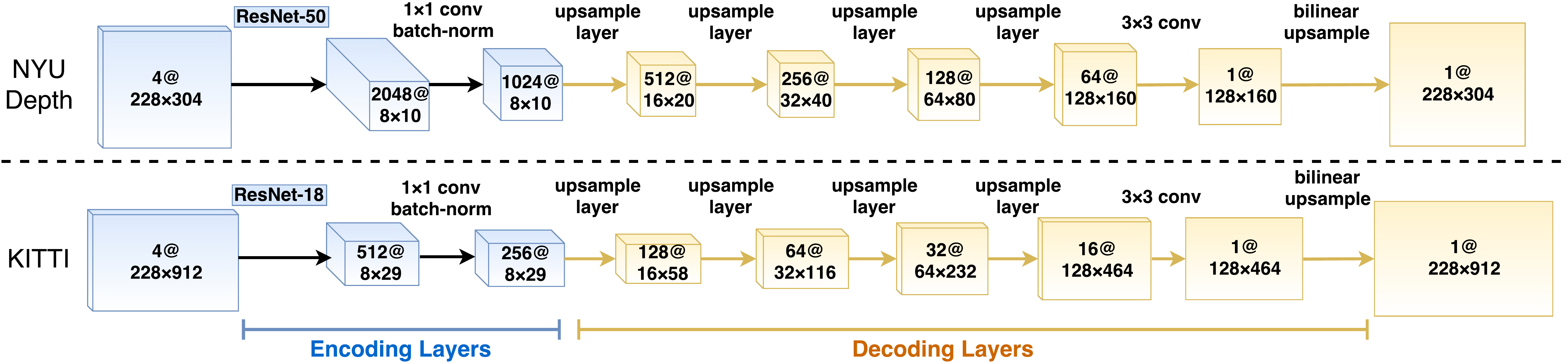} 
\caption{CNN architecture for \nyudepth and \kitti datasets, respectively. Cubes are feature maps, with dimensions represented as \#features@height$\times$width. The encoding layers in blue consist of a ResNet~\cite{he2016deep} and a 3$\times$3 convolution. The decoding layers in yellow are composed of 4 upsampling layers (\upproj) followed by a bilinear upsampling. }
\label{fig:cnn}
\end{figure*}

We found in our experiments that many bottleneck architectures (with an encoder and a decoder) could result in good performance. We chose the final structure based on ~\cite{laina2016deeper} for the sake of benchmarking, because it achieved state-of-the-art accuracy in RGB-based depth prediction. 
The network is tailed to our problem with input data of different modalities, sizes and dimensions. We use two different networks for \kitti and \nyudepth. This is because the \kitti image is triple the size of \nyudepth and consequently the same architecture would require 3 times of GPU memory, exceeding the current hardware capacity. The final structure is illustrated in \prettyref{fig:cnn}. 

The feature extraction (encoding) layers of the network, highlighted in blue, consist of a ResNet~\cite{he2016deep} followed by a convolution layer. More specifically, the ResNet-18 is used for \kitti, and ResNet-50 is used for \nyudepth. The last average pooling layer and linear transformation layer of the original ResNet have been removed. The second component of the encoding structure, the convolution layer, has a kernel size of 3-by-3. 

The decoding layers, highlighted in yellow, are composed of 4 upsampling layers followed by a bilinear upsampling layer. We use the \upproj module proposed by \citet{laina2016deeper} as our upsampling layer, but a deconvolution with larger kernel size can also achieve the same level of accuracy. An empirical comparison of different upsampling layers is shown in \prettyref{sec:results-architecture}. 


\subsection{Depth Sampling}
\label{sec:depth-sampling}
In this section, we introduce the sampling strategy for creating the input sparse depth image from the ground truth. 

During training, the input sparse depth $\inputD$ is sampled randomly from the ground truth depth image $\gtD$ on the fly. In particular, for any targeted number of depth samples $m$ (fixed during training), we compute a Bernoulli probability $p = \frac m n$, where $n$ is the total number of valid depth pixels in $\gtD$. Then, for any pixel $(i,j)$, 
\begin{align}
\inputD(i,j) =
  \begin{cases}
      \gtD(i,j),& \text{with probability } p\\
      0, & \text{otherwise}
  \end{cases}
\end{align}

With this sampling strategy, the actual number of non-zero depth pixels varies for each training sample around the expectation $m$. Note that this sampling strategy 
is different from dropout~\cite{srivastava2014dropout}, which scales up the output by $1/p$ during training to compensate for deactivated neurons. The purpose of our sampling strategy is to increase robustness of the network against different number of inputs and to create more training data (\ie, a data augmentation technique). It is worth exploring how injection of random noise and a different sampling strategy (\eg, feature points) would affect the performance of the network.

\subsection{Data Augmentation}
\label{sec:method-augmentation}
We augment the training data in an online manner with random transformations, including
\begin{itemize}
  \item {\em Scale}: color images are scaled by a random number $s \in [1, 1.5]$, and depths are divided by $s$.
  \item {\em Rotation}: color and depths are both rotated with a random degree $r \in [-5,5]$.
  \item {\em Color Jitter}: the brightness, contrast, and saturation of color images are each scaled by $k_i \in [0.6,1.4]$. 
  \item {\em Color Normalization}: RGB is normalized through mean subtraction and division by standard deviation.
  \item {\em Flips}: color and depths are both horizontally flipped with a 50\% chance.
\end{itemize}

Nearest neighbor interpolation, rather than the more common bi-linear or bi-cubic interpolation, is used in both scaling and rotation to avoid creating spurious sparse depth points. We take the center crop from the augmented image so that the input size to the network is consistent. 

\subsection{Loss Function}
One common and default choice of loss function for regression problems is the mean squared error (\ltwo). \ltwo is sensitive to outliers in the training data since it penalizes more heavily on larger errors. During our experiments we found that the \ltwo loss function also yields visually undesirable, over-smooth boundaries instead of sharp transitions. 

Another common choice is the Reversed Huber (denoted as \berhu) loss function~\cite{owen2007robust}, defined as
\begin{align}
\berhumath(e) = 
  \begin{cases}
      \abs{e},& \text{if } \abs{e} \leq c\\
      \frac {e^2 + c^2} {2c}, & \text{otherwise}
  \end{cases}
\end{align}
\cite{laina2016deeper} uses a batch-dependent parameter $c$, computed as 20\% of the maximum absolute error over all pixels in a batch. Intuitively, \berhu acts as the mean absolute error (\lone) when the element-wise error falls below $c$, and behaves approximately as \ltwo when the error exceeds $c$. 

In our experiments, besides the aforementioned two loss functions, we also tested \lone and found that it produced slightly better results on the RGB-based depth prediction problem. The empirical comparison is shown in \prettyref{sec:results-architecture}. As a result, we use \lone as our default choice throughout the paper for its simplicity and performance.


\section{Experiments}
We implement the network using Torch~\cite{torch}. Our models are trained on the \nyudepth and \kitti odometry datasets using a NVIDIA Tesla P100 GPU with 16GB memory. The weights of the ResNet in the encoding layers (except for the first layer which has different number of input channels) are initialized with models pretrained on the ImageNet dataset~\cite{russakovsky2015imagenet}.
We use a small batch size of 16 and train for 20 epochs. The learning rate starts at 0.01, and is reduced to 20\% every 5 epochs. A small weight decay of $10^{-4}$ is applied for regularization. 

\subsection{The \nyudepth Dataset}
The \nyudepth dataset~\cite{silberman2012indoor} consists of RGB and depth images collected from 464 different indoor scenes with a Microsoft Kinect. We use the official split of data, where 249 scenes are used for training and the remaining 215 for testing. In particular, for the sake of benchmarking, the small labeled test dataset with 654 images is used for evaluating the final performance, as seen in previous work~\cite{laina2016deeper,eigen2014depth}. 

For training, we sample spatially evenly from each raw video sequence from the training dataset, generating roughly 48k synchronized depth-RGB image pairs. The depth values are projected onto the RGB image and in-painted with a cross-bilateral filter using the official toolbox. Following~\cite{laina2016deeper,eigen2014depth}, the original frames of size 640$\times$480 are first down-sampled to half and then center-cropped, producing a final size of 304$\times$228. 


\subsection{The \kitti Odometry Dataset}
In this work we use the {\em odometry} dataset, which includes both camera and LiDAR measurements
. The {\em odometry} dataset consists of 22 sequences. Among them, one half is used for training while the other half is for evaluation. We use all 46k images from the training sequences for training the neural network, and a random subset of 3200 images from the test sequences for the final evaluation.

We use both left and right RGB cameras as unassociated shots. The Velodyne LiDAR measurements are projected onto the RGB images. Only the bottom crop (912$\times$228) is used, since the LiDAR returns no measurement to the upper part of the images. Compared with \nyudepth, even the ground truth is sparse for \kitti, typically with only 18k projected measurements out of the 208k image pixels. 


\subsection{Error Metrics}
We evaluate each method using the following metrics:
\begin{itemize}
  \item \rmse: root mean squared error
  \item \absrel: mean absolute relative error
  \item $\delta_i$: percentage of predicted pixels where the relative error is within a threshold. Specifically, 
  $$\delta_i = \frac {\cart{ \left\{\hat{y}_i: \max \left\{ \frac {\hat{y}_i} {y_i}, \frac {y_i} {\hat{y}_i} \right\}<1.25^i \right\}}} {\cart{\{y_i\}}},$$
  where $y_i$ and $\hat{y}_i$ are respectively the ground truth and the prediction, and $\mathbf{card}$ is the cardinality of a set. A higher $\delta_i$ indicates better prediction.
\end{itemize}


\section{Results}
\label{sec:results}
In this section we present all experimental results. First, we evaluate the performance of our proposed method with different loss functions and network components on the prediction accuracy in \prettyref{sec:results-architecture}. Second, we compare the proposed method with state-of-the-art methods on both the \nyudepth and the \kitti datasets in \prettyref{sec:results-literature}. Third, In \prettyref{sec:results-samples}, we explore the impact of number of sparse depth samples on the performance. Finally, in \prettyref{sec:results-dense} and \prettyref{sec:results-lidar}, we demonstrate two use cases of our proposed algorithm in creating dense maps and LiDAR super-resolution.
\subsection{Architecture Evaluation}
\label{sec:results-architecture}
In this section we present an empirical study on the impact of different loss functions and network components
on the depth prediction accuracy. The results are listed in \prettyref{tab:architecture}.


\begin{table}[htbp]
\centering
\footnotesize
\setlength\tabcolsep{1.5pt} 

\begin{tabular}{| c || c c c || *{2}{ c } | *{3}{ c } | }
\hline
Problem & Loss & Encoder & Decoder & \rmse & \absrel & \deltas{1} & \deltas{2} & \deltas{3}\\ \hline\hline 

\RGB 
& \ltwo & \conv & \deconv{2} & 0.610 & 0.185  & 71.8 & 93.4 & 98.3 \\
& \berhu & \conv & \deconv{2} & 0.554 & 0.163 & \bf{77.5} & 94.8 & \bf{98.7} \\
& \textbf{\lone} & \conv & \deconv{2} & \bf{0.552} & \bf{0.159} & \bf{77.5} & \bf{95.0} & \bf{98.7} \\ 
\hdashline
& & \conv & \deconv{3} & 0.533 & 0.151 & 79.0 & 95.4 & 98.8 \\ 
& & \conv & \upconv & 0.529 & 0.149 & 79.4 & \bf{95.5} & \bf{98.9} \\
& & \conv & \upproj & \bf{0.528} & \bf{0.144} & \bf{80.3} & 95.2 & 98.7\\
\hline \hline

\RGBd
& \lone & \channeldrop & \upproj & 0.361 & 0.105 & 90.8 & 98.4 & 99.6 \\ 
& & \depthwise & \upproj & \bf{0.261} & 0.054 & \bf{96.2} & \bf{99.2} & 99.7 \\ 
& & \conv & \upproj & 0.264 & \bf{0.053} & 96.1 & \bf{99.2} & \bf{99.8} \\

\hline
\end{tabular}

\caption{Evaluation of loss functions, upsampling layers and the first convolution layer. \RGBd has an average sparse depth input of 100 samples. (a) comparison of loss functions is listed in Row 1 - 3; (b) comparison of upsampling layers is in Row 2 - 4; (c) comparison of the first convolution layers is in the 3 bottom rows.}
\label{tab:architecture}
\end{table}

\subsubsection{Loss Functions}
To compare the loss functions we use the same network architecture, where the upsampling layers are simple deconvolution with a 2$\times$2 kernel (denoted as \deconv{2}). \ltwo, \berhu and \lone loss functions are listed in the first three rows in \prettyref{tab:architecture} for comparison. As shown in the table, both \berhu and \lone significantly outperform \ltwo. In addition, \lone produces slightly better results than \berhu. Therefore, we use \lone as our default choice of loss function.

\subsubsection{Upsampling Layers}
We perform an empirical evaluation of different upsampling layers, including deconvolution with kernels of different sizes (\deconv{2} and \deconv{3}), as well as the \upconv and \upproj modules proposed by \citet{laina2016deeper}. The results are listed from row 3 to 6 in \prettyref{tab:architecture}.

We make several observations. Firstly, deconvolution with a 3$\times$3 kernel (\ie, \deconv{3}) outperforms the same component with only a 2$\times$2 kernel (\ie, \deconv{2}) in every single metric. Secondly, since both \deconv{3} and \upconv have a receptive field of 3$\times$3 (meaning each output neuron is computed from a neighborhood of 9 input neurons), they have comparable performance. Thirdly, with an even larger receptive field of 4$\times$4, the \upproj module outperforms the others. We choose to use \upproj as a default choice.

\subsubsection{First Convolution Layer}
Since our \RGBd input data comes from different sensing modalities, its 4 input channels (R, G, B, and depth) have vastly different distributions and support. 
We perform a simple analysis on the first convolution layer and explore three different options. 

The first option is the regular spatial convolution (\conv). The second option is depthwise separable convolution (denoted as \depthwise), which consists of a spatial convolution performed independently on each input channel, followed by a pointwise convolution across different channels with a window size of 1. The third choice is channel dropout (denoted as \channeldrop), through which each input channel is preserved as is with some probability $p$, and zeroed out with probability $1-p$.

The bottom 3 rows compare the results from the 3 options. The networks are trained using \RGBd input with an average of 100 sparse input samples. \depthwise and \conv yield very similar results, and both significantly outperform the \channeldrop layer. Since the difference is small, for the sake of comparison consistency, we will use the convolution layer for all experiments.


\subsection{Comparison with the State-of-the-Art}
\label{sec:results-literature}
In this section, we compare with existing methods. 

\subsubsection{\nyudepth Dataset}
We compare with RGB-based approaches~\cite{roy2016monocular,eigen2014depth,laina2016deeper}, as well as the fusion approach~\cite{liao2017parse} that utilizes an additional 2D laser scanner mounted on a ground robot. The quantitative results are listed in \prettyref{tab:methods-nyu}. 


\begin{table}[htbp]
\centering
\footnotesize
\setlength\tabcolsep{3pt} 

\begin{tabular}{| c | c || c | *{2}{ c } | *{3}{ c } | }
\hline
Problem & \#Samples & Method & \rmse & \absrel & \deltas{1} & \deltas{2} & \deltas{3}\\ \hline\hline 

\RGB 
& 0 & Roy~\etal~\cite{roy2016monocular} & 0.744 & 0.187 & - & - & - \\ 
& 0 & Eigen~\etal~\cite{eigen2015predicting} & 0.641 & 0.158 & 76.9 & 95.0 & 98.8 \\
& 0 & Laina~\etal~\cite{laina2016deeper} & 0.573 & \bf{0.127} & \bf{81.1} & 95.3 & 98.8 \\
& 0 & Ours-\RGB & \bf{0.514} & 0.143 & 81.0 & \bf{95.9} & \bf{98.9} \\ 
\hline \hline

\sd
& 20 & Ours-\sd & 0.461 & 0.110 & 87.2 & 96.1 & 98.8 \\  
& 50 & Ours-\sd & 0.347 & 0.076 & 92.8 & 98.2 & 99.5 \\  
& 200 & Ours-\sd & \bf0.259 & \bf0.054 & \bf96.3 & \bf99.2 & \bf99.8 \\  
\hline \hline

\RGBd 
& 225 & Liao~\etal~\cite{liao2017parse} & 0.442 & 0.104 & 87.8 & 96.4 & 98.9 \\
& 20 & Ours-\RGBd & 0.351 & 0.078 & 92.8 & 98.4 & 99.6  \\
& 50 & Ours-\RGBd & 0.281 & 0.059 & 95.5 & 99.0 & 99.7 \\
& 200 & Ours-\RGBd & \bf0.230 & \bf0.044 & \bf97.1 & \bf99.4 & \bf99.8 \\
\hline
\end{tabular}

\caption{Comparison with state-of-the-art on the \nyudepth dataset. The values are those originally reported by the authors in their respective paper}

\label{tab:methods-nyu}
\end{table}

\begin{figure}[htbp]
\centering
\begin{minipage}{\textwidth}
\newcommand{\figWidth}{ 0.305\linewidth } 
\def\arraystretch{5.8}    
\setlength\tabcolsep{0.3mm} 
\scriptsize
\begin{tabular}{ p{0.3cm} c c c }
  \begin{minipage}[c]{\figWidth}\centering
  (a) 
  \end{minipage}
  &
  \begin{minipage}[c]{\figWidth}\centering
  \includegraphics[width=\linewidth]{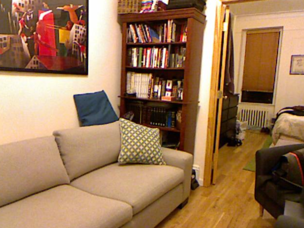} 
  \end{minipage}
  & 
  \begin{minipage}[c]{\figWidth}\centering
  \includegraphics[width=\linewidth]{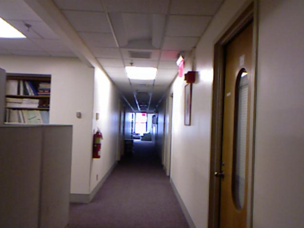} 
  \end{minipage}
  & 
  \begin{minipage}[c]{\figWidth}\centering
  \includegraphics[width=\linewidth]{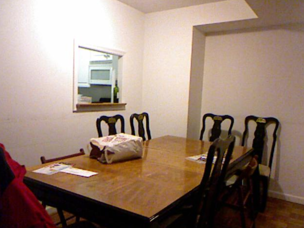} 
  \end{minipage}
  \\
  \begin{minipage}[c]{\figWidth}\centering
  (b)
  \end{minipage}
  &
  \begin{minipage}[c]{\figWidth}\centering
  \includegraphics[width=\linewidth]{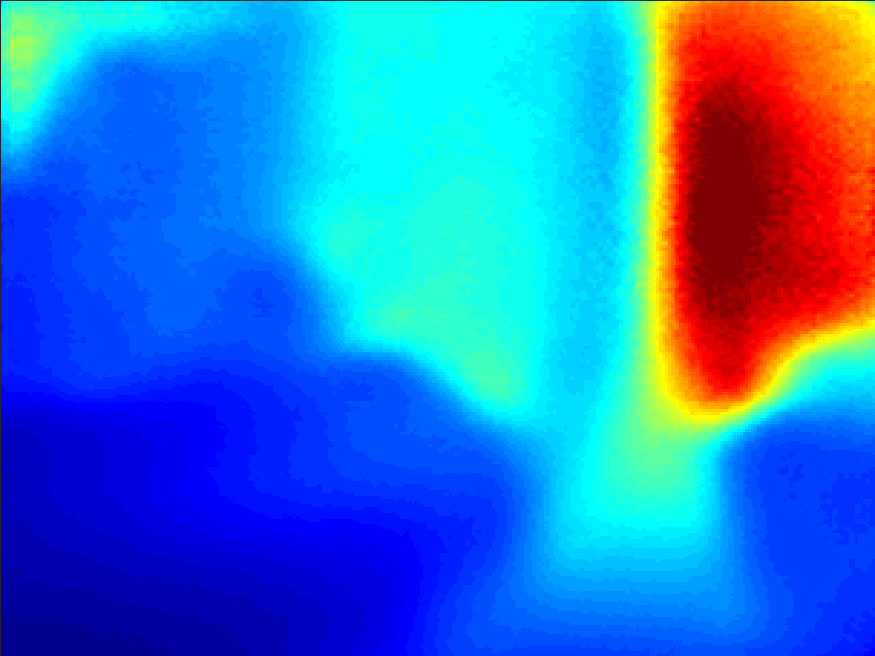} 
  \end{minipage}
  & 
  \begin{minipage}[c]{\figWidth}\centering
  \includegraphics[width=\linewidth]{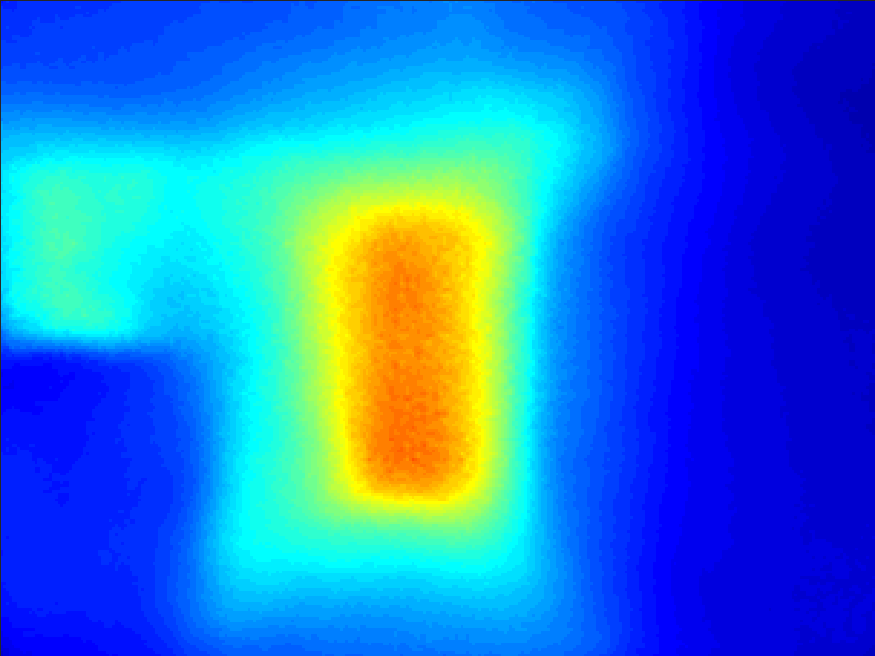} 
  \end{minipage}
  & 
  \begin{minipage}[c]{\figWidth}\centering
  \includegraphics[width=\linewidth]{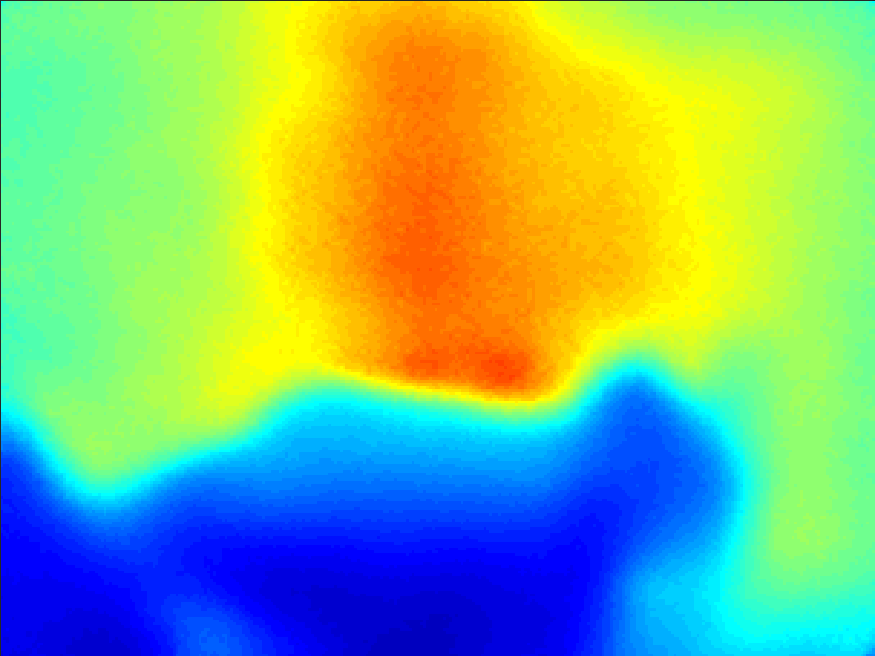} 
  \end{minipage}
  \\ 
  \begin{minipage}[c]{\figWidth}\centering
  (c)
  \end{minipage}
  &
  \begin{minipage}[c]{\figWidth}\centering
  \includegraphics[width=\linewidth]{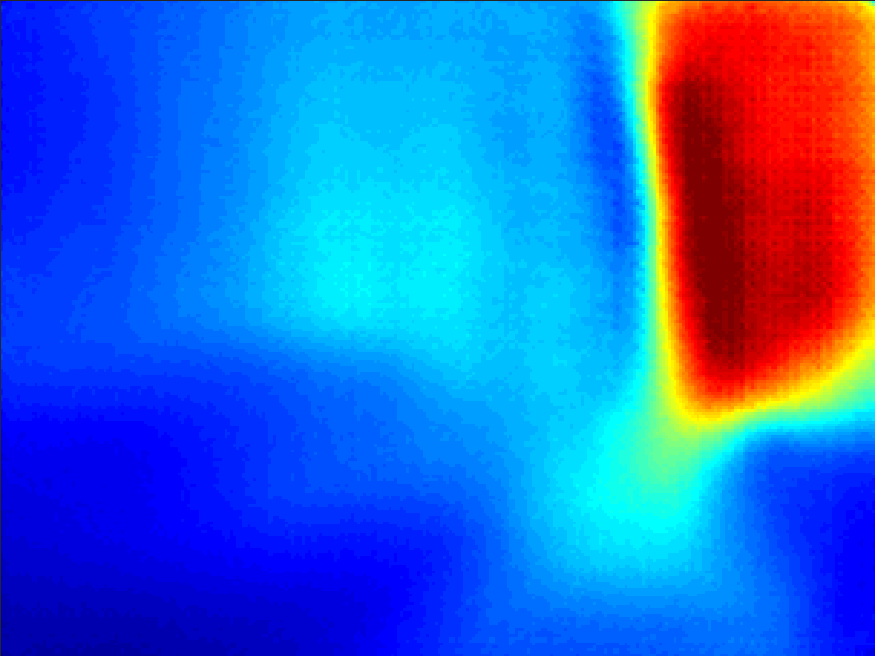} 
  \end{minipage}
  & 
  \begin{minipage}[c]{\figWidth}\centering
  \includegraphics[width=\linewidth]{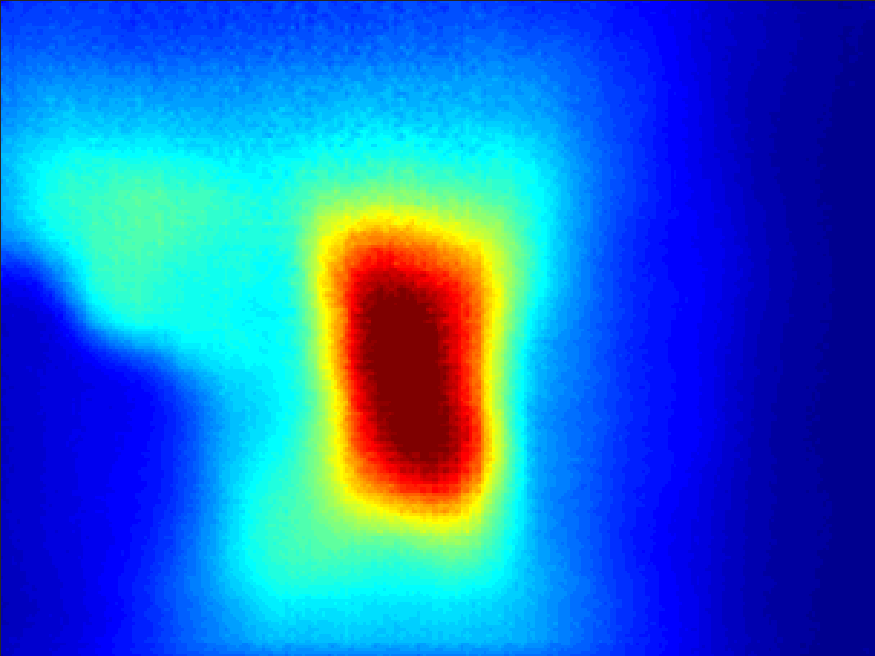} 
  \end{minipage}
  & 
  \begin{minipage}[c]{\figWidth}\centering
  \includegraphics[width=\linewidth]{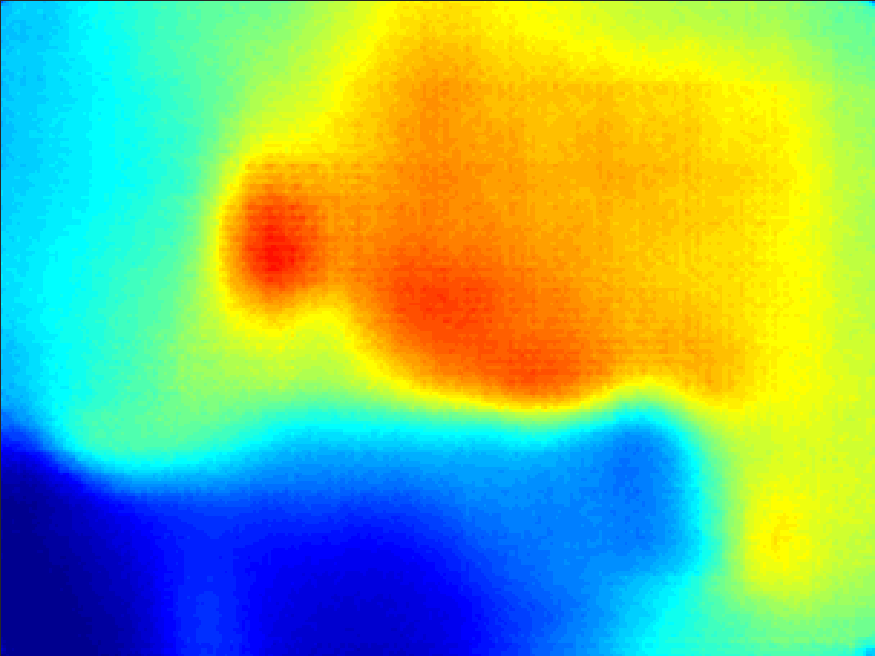}  
  \end{minipage}
  \\
  \begin{minipage}[c]{\figWidth}\centering
  (d)
  \end{minipage}
  &
  \begin{minipage}[c]{\figWidth}\centering
  \includegraphics[width=\linewidth]{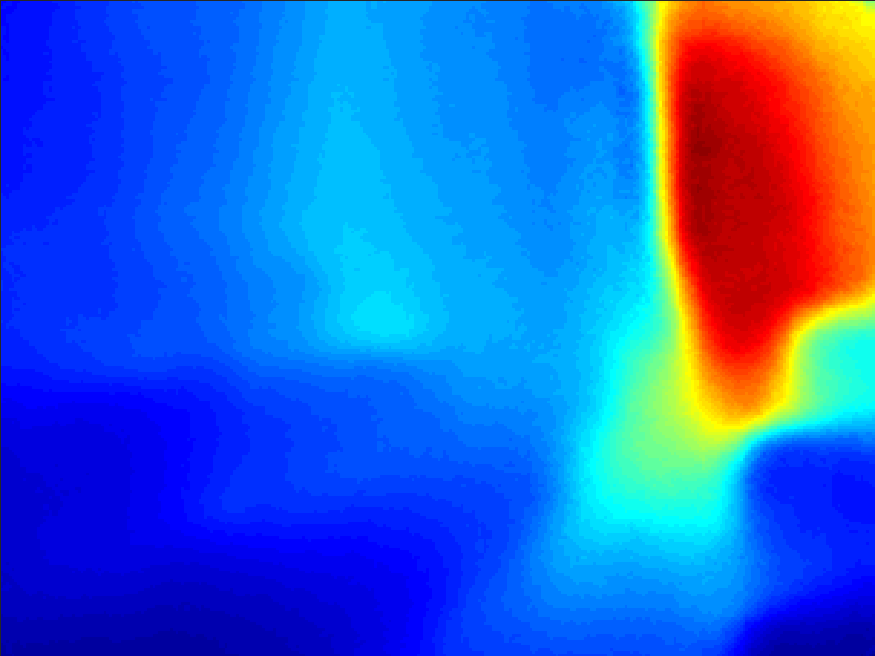} 
  \end{minipage}
  & 
  \begin{minipage}[c]{\figWidth}\centering
  \includegraphics[width=\linewidth]{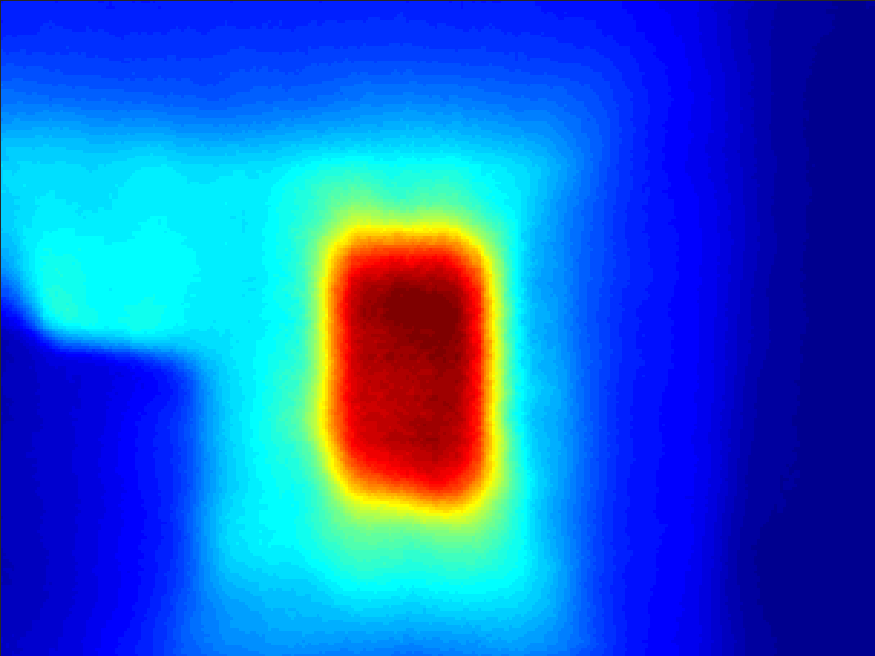} 
  \end{minipage}
  & 
  \begin{minipage}[c]{\figWidth}\centering
  \includegraphics[width=\linewidth]{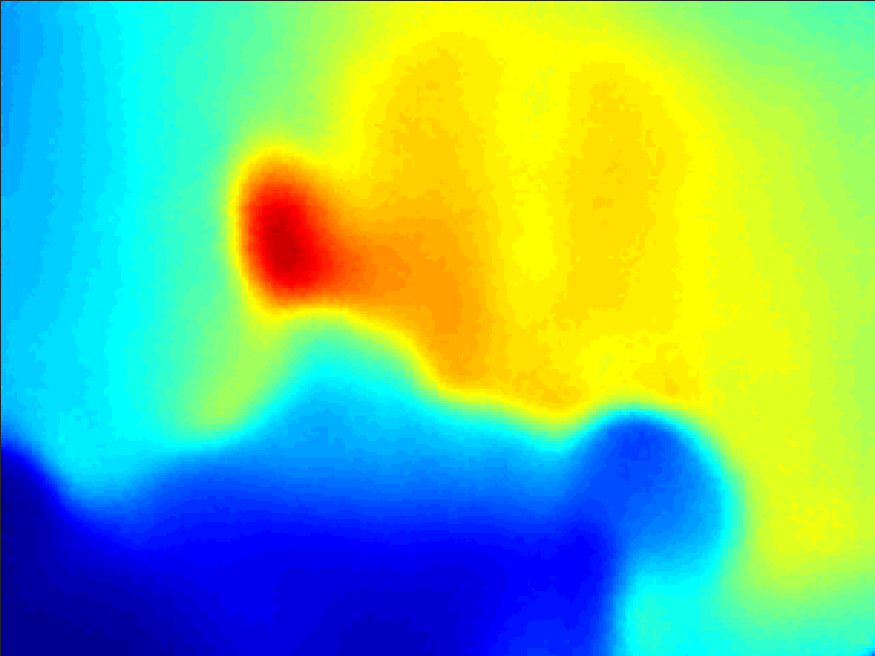}  
  \end{minipage}
  \\
  \begin{minipage}[c]{\figWidth}\centering
  (e)
  \end{minipage}
  &
  \begin{minipage}[c]{\figWidth}\centering
  \includegraphics[width=\linewidth]{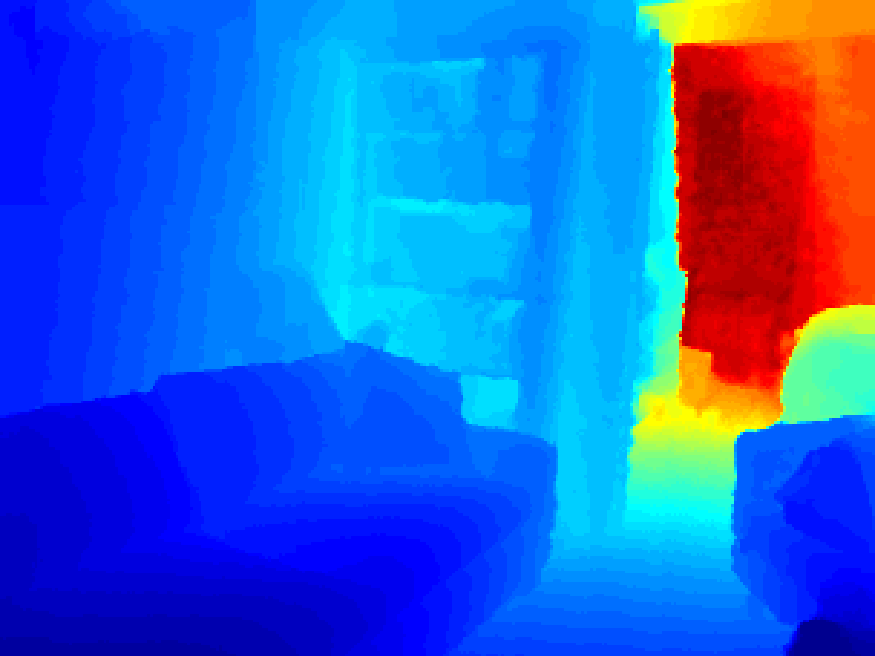} 
  \end{minipage}
  & 
  \begin{minipage}[c]{\figWidth}\centering
  \includegraphics[width=\linewidth]{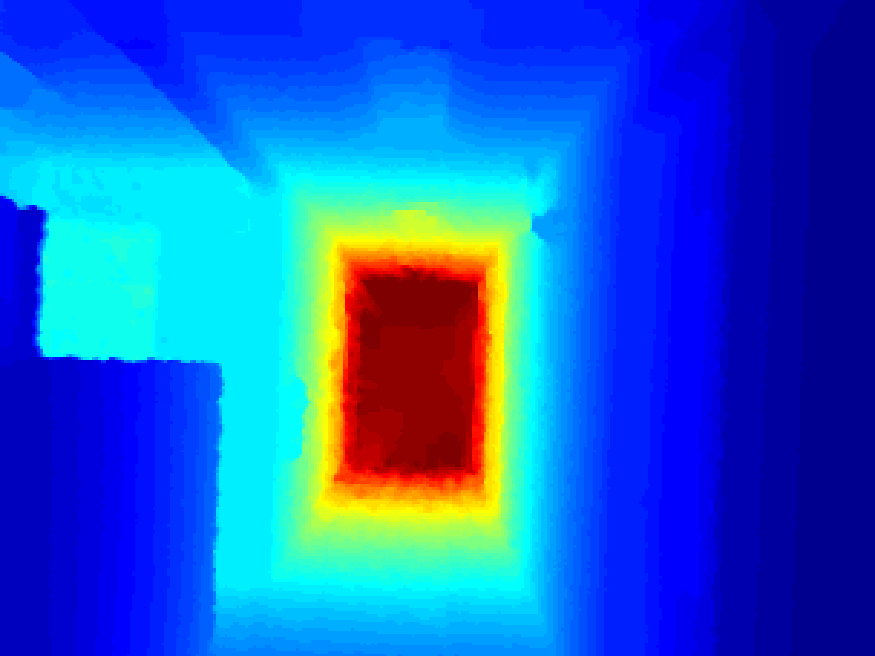} 
  \end{minipage}
  & 
  \begin{minipage}[c]{\figWidth}\centering
  \includegraphics[width=\linewidth]{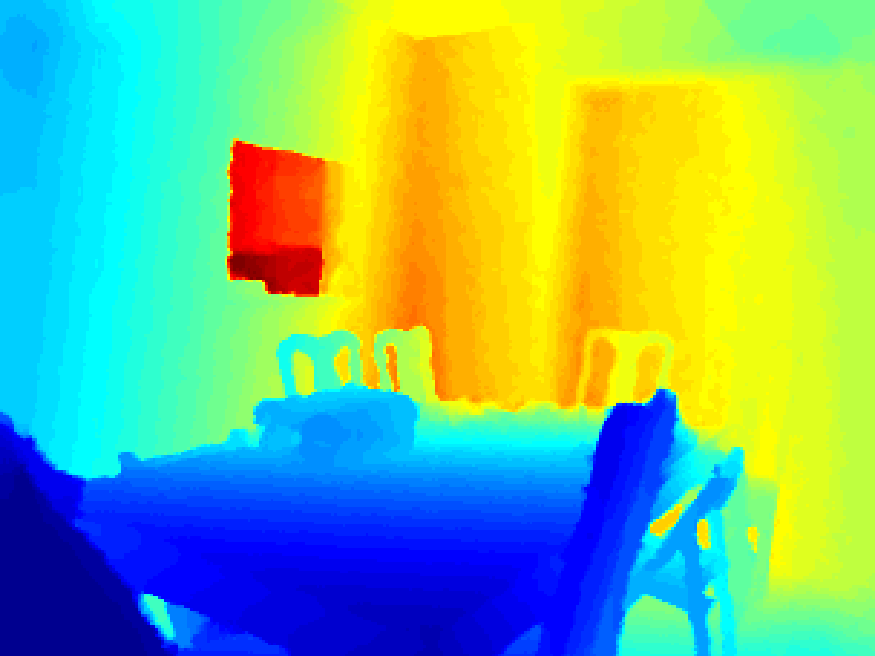}  
  \end{minipage}
\end{tabular}
\end{minipage}

\caption{Predictions on \nyudepth. From top to bottom: (a) rgb images; (b) RGB-based prediction; (c) \sd prediction with 200  and no rgb; (d) \RGBd prediction with 200 sparse depth and rgb; (e) ground truth depth.}
\label{fig:examples-nyu}
\end{figure}

Our first observation from Row 2 and Row 3 is that, with the same network architecture, we can achieve a slightly better result (albeit higher \absrel) by replacing the \berhu loss function proposed in~\cite{laina2016deeper} with a simple \lone. Secondly, by comparing problem group \RGB (Row 3) and problem group \sd (\eg, Row 4), we draw the conclusion that an extremely small set of 20 sparse depth samples (without color information) already produces significantly better predictions than using \RGB. Thirdly, by comparing problem group \sd and proble group \RGBd row by row with the same number of samples, it is clear that the color information does help improve the prediction accuracy. In other words, our proposed method is able to learn a suitable representation from both the RGB images and the sparse depth images. Finally, we compare against \cite{liao2017parse} (bottom row). Our proposed method, even using only 100 samples, outperforms \cite{liao2017parse} with 225 laser measurements. This is because our samples are spatially uniform, and thus provides more information than a line measurement. A few examples of our predictions with different inputs are displayed in \prettyref{fig:examples-nyu}.

\subsubsection{KITTI Dataset}

\begin{table}[htbp]
\centering
\footnotesize
\setlength\tabcolsep{2pt} 

\begin{tabular}{| c | c || c | *{2}{ c } | *{3}{ c } | }
\hline
Problem & \#Samples & Method & \rmse & \absrel & \deltas{1} & \deltas{2} & \deltas{3}\\ \hline\hline 

\RGB 
& 0 & Make3D~\cite{laina2016deeper} & 8.734 & 0.280 & 60.1 & 82.0 & 92.6 \\
& 0 & Mancini~\cite{mancini2016fast} & 7.508 & - & 31.8 & 61.7 & 81.3 \\ 
& 0 & \citet{eigen2014depth} & 7.156 & \bf{0.190} & \bf{69.2} & 89.9 & \bf{96.7} \\
& 0 & Ours-\RGB & \bf6.266 & 0.208 & 59.1 & \bf90.0 & 96.2 \\ 
\hline \hline


\RGBd 
& $\sim$650 & full-MAE~\cite{cadena2016multi} & 7.14 & 0.179 & 70.9 & 88.8 & 95.6 \\
& 50 & Ours-\RGBd & 4.884 & 0.109 & 87.1 & 95.2 & 97.9 \\
& 225 & Liao~\etal~\cite{liao2017parse} & 4.50 & 0.113 & 87.4 & 96.0 & 98.4 \\
& 100 & Ours-\RGBd & 4.303 & 0.095 & 90.0 & 96.3 & 98.3 \\
& 200 & Ours-\RGBd & 3.851 & 0.083 & 91.9 & 97.0 & 98.6 \\
& 500 & Ours-\RGBd & \bf3.378 & \bf0.073 & \bf93.5 & \bf97.6 & \bf98.9 \\
\hline
\end{tabular}

\caption{Comparison with state-of-the-art on the \kitti dataset. The Make3D values are those reported in \cite{eigen2014depth}}

\label{tab:methods-kitti}
\end{table}
The \kitti dataset is more challenging for depth prediction, since the maximum distance is 100 meters as opposed to only 10 meters in the \nyudepth dataset. A greater performance boost can be obtained from using our approach. Although the training and test data are not the same across different methods, the scenes are similar in the sense that they all come from the same sensor setup on a car and the data were collected during driving. We report the values from each work in \prettyref{tab:methods-kitti}.


\begin{figure}[hbtp]
\centering
\begin{minipage}{\textwidth}
\newcommand{\figWidth}{ 0.915\linewidth } 
\newcommand{\figHeight}{ 0.2\linewidth } 
\def\arraystretch{4.6}
\setlength\tabcolsep{0.3mm} 
\scriptsize
\begin{tabular}{ p{0.7cm} c }
  \begin{minipage}[c]{\figWidth}\centering
    (a) \\
    color
  \end{minipage}
  &
  \begin{minipage}[c]{\figWidth}\centering
  \includegraphics[width=\linewidth, height=\figHeight]{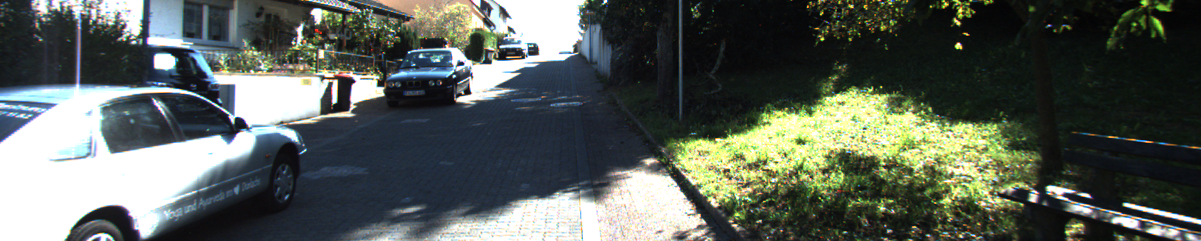} 
  \end{minipage}
  \\ 
  \begin{minipage}[c]{\figWidth}\centering
    (b) \\
    sparse \\
    input \\
    depth
  \end{minipage}
  &
  \begin{minipage}[c]{\figWidth}\centering
  \includegraphics[width=\linewidth, height=\figHeight]{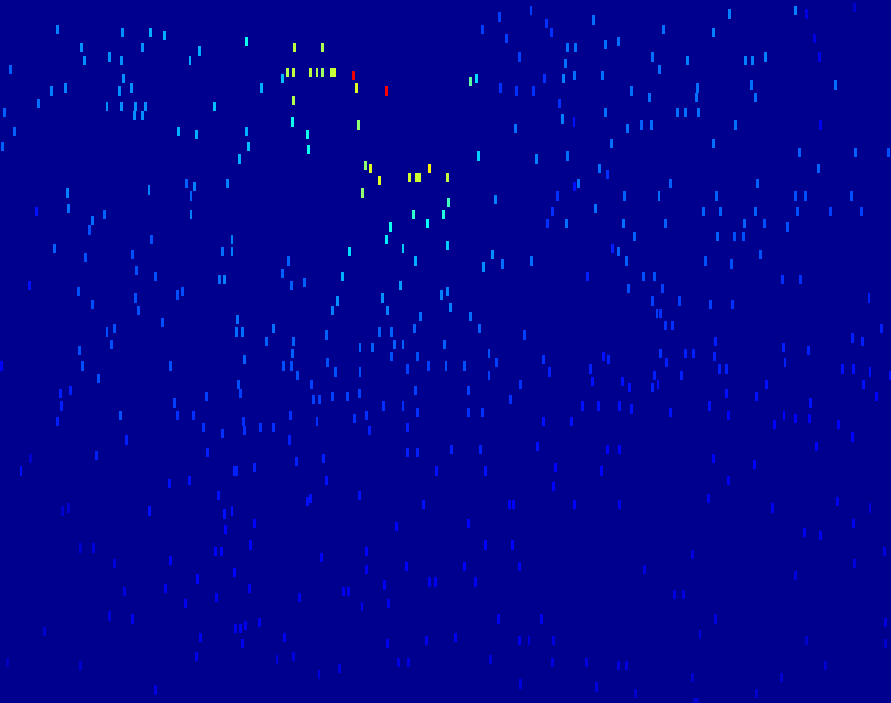} 
  \end{minipage}
  \\
  \begin{minipage}[c]{\figWidth}\centering
    (c) \\
    \RGBd-500
  \end{minipage}
  &
  \begin{minipage}[c]{\figWidth}\centering
  \includegraphics[width=\linewidth, height=\figHeight]{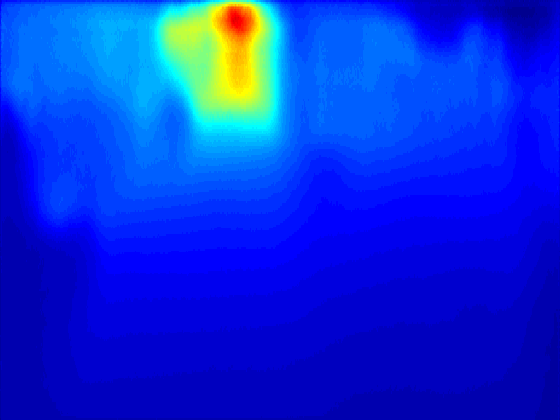}   
  \end{minipage}
  \\
  \begin{minipage}[c]{\figWidth}\centering
    (d) \\
    ground \\
    truth
  \end{minipage}
  &
  \begin{minipage}[c]{\figWidth}\centering
  \includegraphics[width=\linewidth, height=\figHeight]{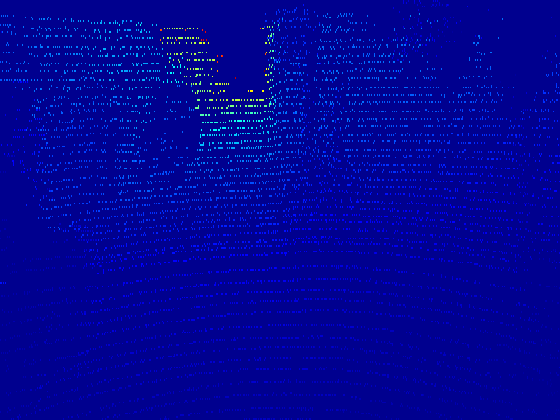} 
  \end{minipage}
\end{tabular}
\end{minipage}

\caption{Example of prediction on \kitti. From top to bottom: (a) RGB; (b) sparse depth; (c) \RGBd dense prediction; (d) ground truth depth projected from LiDAR.}
\label{fig:examples-kitti}
\end{figure}
The results in the first \RGB group demonstrate that \RGB-based depth prediction methods fail in outdoor scenarios, with a pixel-wise \rmse of close to 7 meters. Note that we use sparsely labeled depth image projected from LiDAR, instead of dense disparity maps computed from stereo cameras as in \cite{eigen2014depth}. In other words, we have a much smaller training dataset compared with~\cite{mancini2016fast,eigen2014depth}. 

An additional 500 depth samples bring the \rmse to 3.3 meters, a half of the \RGB approach, and boosts \deltas{1} from only 59.1\% to 93.5\%. Our performance also compares favorably to other fusion techniques including \cite{liao2017parse,cadena2016multi}, and at the same time demands fewer samples.



\subsection{On Number of Depth Samples}
\label{sec:results-samples}
In this section, we explore the relation between the prediction accuracy and the number of available depth samples. We train a network for each different input size for optimal performance. We compare the performance for all three kinds of input data, including \RGB, \sd, and \RGBd. The performance of RGB-based depth prediction is independent of input sample size and is thus plotted as a horizontal line for benchmarking. 

\begin{figure}[htbp]
\centering
\includegraphics[width=\linewidth]{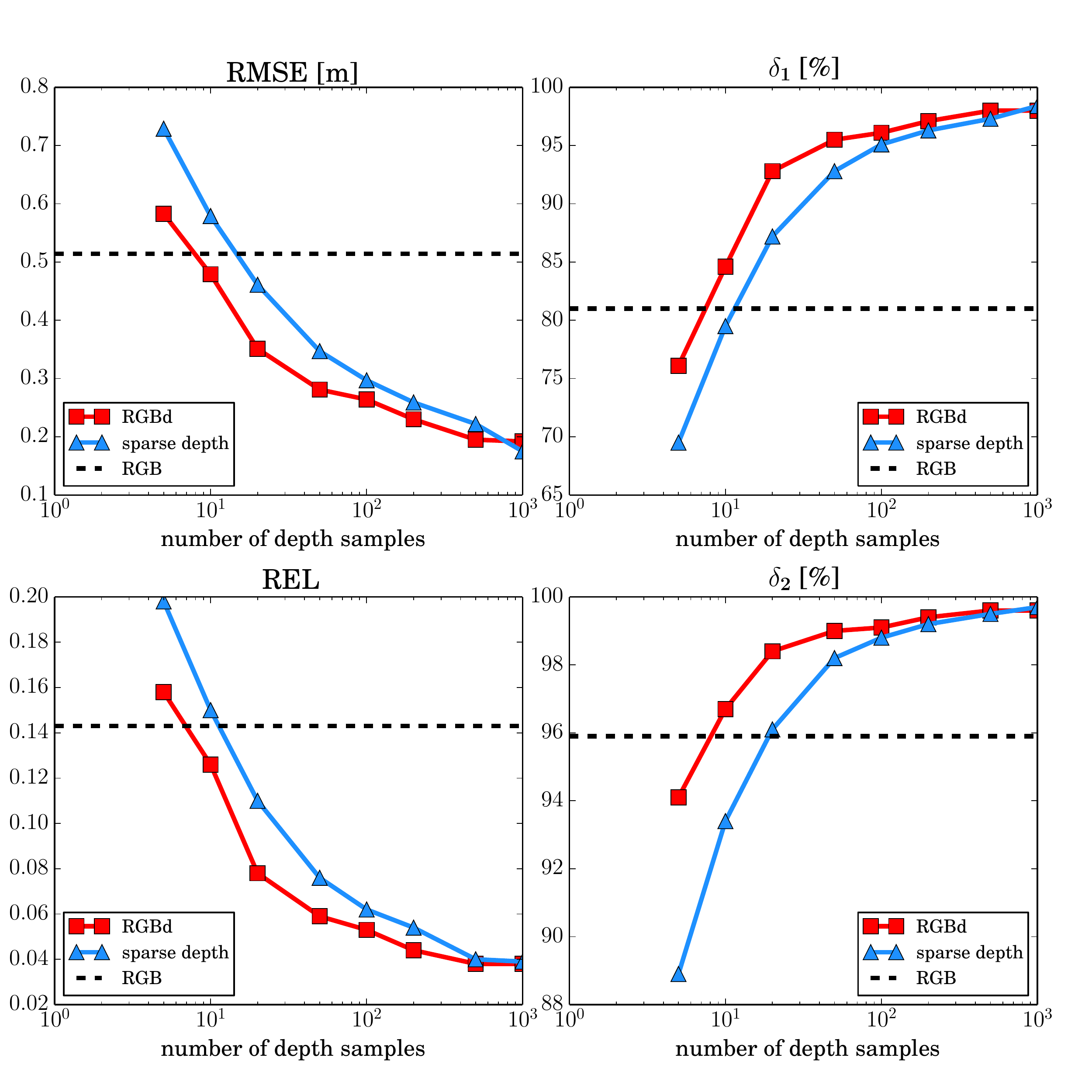} 
\caption{Impact of number of depth sample on the prediction accuracy on the \nyudepth dataset. Left column: lower is better; right column: higher is better.}
\label{fig:acc_vs_samples_nyu}
\end{figure}

\begin{figure}[htbp]
\centering
\includegraphics[width=\linewidth]{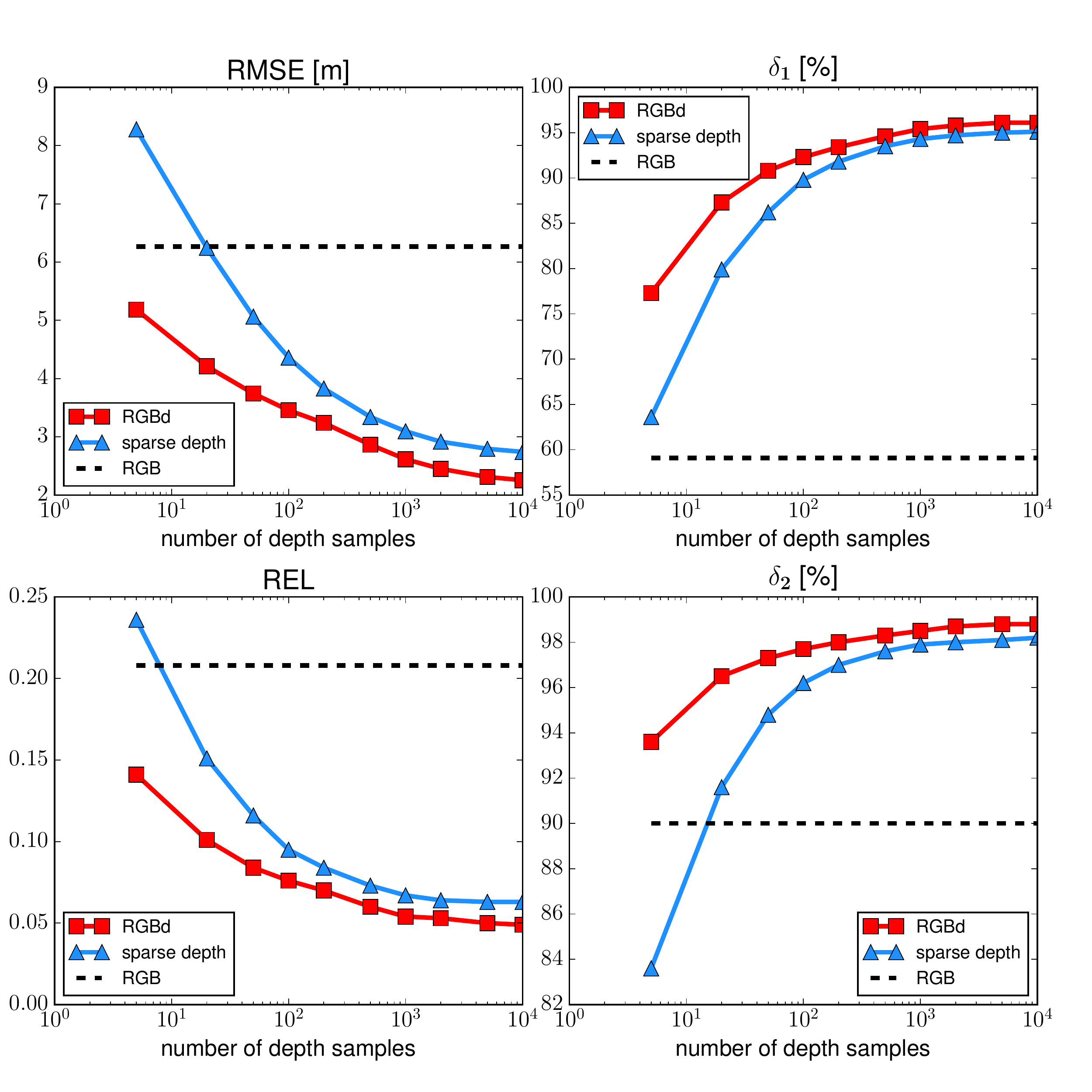} 
\caption{Impact of number of depth sample on the prediction accuracy on the \kitti dataset. Left column: lower is better; right column: higher is better.}
\label{fig:acc_vs_samples_kitti}
\end{figure}

On the \nyudepth dataset in \prettyref{fig:acc_vs_samples_nyu}, the \RGBd outperforms \RGB with over 10 depth samples and the performance gap quickly increases with the number of samples. With a set of 100 samples, the \rmse of \RGBd decreases to around 25cm, half of \RGB (51cm). The \absrel sees a larger improvement (from 0.15 to 0.05, reduced by two thirds). On one hand, the \RGBd approach consistently outperforms \sd, which indicates that the learned model is indeed able to extract information not only from the sparse samples alone, but also from the colors. On the other hand, the performance gap between \RGBd and \sd shrinks as the sample size increases. Both approaches perform equally well when sample size goes up to 1000, which accounts for less than 1.5\% of the image pixels and is still a small number compared with the image size. This observation indicates that the information extracted from the sparse sample set dominates the prediction when the sample size is sufficiently large, and in this case the color cue becomes almost irrelevant.

The performance gain on the \kitti dataset is almost identical to \nyudepth, as shown in \prettyref{fig:acc_vs_samples_kitti}. With 100 samples the \rmse of \RGBd decreases from 7 meters to a half, 3.5 meters. This is the same percentage of improvement as on the \nyudepth dataset. Similarly, the \absrel is reduced from 0.21 to 0.07, again the same percentage of improvement as the \nyudepth.

On both datasets, the accuracy saturates as the number of depth samples increases. Additionally, the prediction has blurry boundaries even with many depth samples (see \prettyref{fig:super-resolution}). We believe both phenomena can be attributed to the fact that fine details are lost in bottleneck network architectures. It remains further study if additional skip connections from encoders to decoders help improve performance.

\subsection{Application: Dense Map from Visual Odometry Features}
\label{sec:results-dense}
In this section, we demonstrate a use case of our proposed method in sparse visual SLAM and visual inertial odometry (VIO). The best-performing algorithms for SLAM and VIO are usually sparse methods, which represent the environment with sparse 3D landmarks. 
Although sparse SLAM/VIO algorithms are robust and efficient, the output map is in the form of sparse point clouds and is not useful for other applications (\eg motion planning). 


\begin{figure}[htbp]
\centering
\begin{minipage}{\textwidth}
\newcommand{\figWidth}{ 0.49\linewidth } 
\setlength\tabcolsep{0.3mm} 
\begin{tabular}{ c c }
  \begin{minipage}[b]{\figWidth}\centering
  \includegraphics[width=\linewidth]{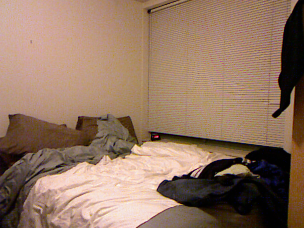} \\
  (a) RGB
  \end{minipage}
  &
  \begin{minipage}[b]{\figWidth}\centering
  \includegraphics[width=\linewidth]{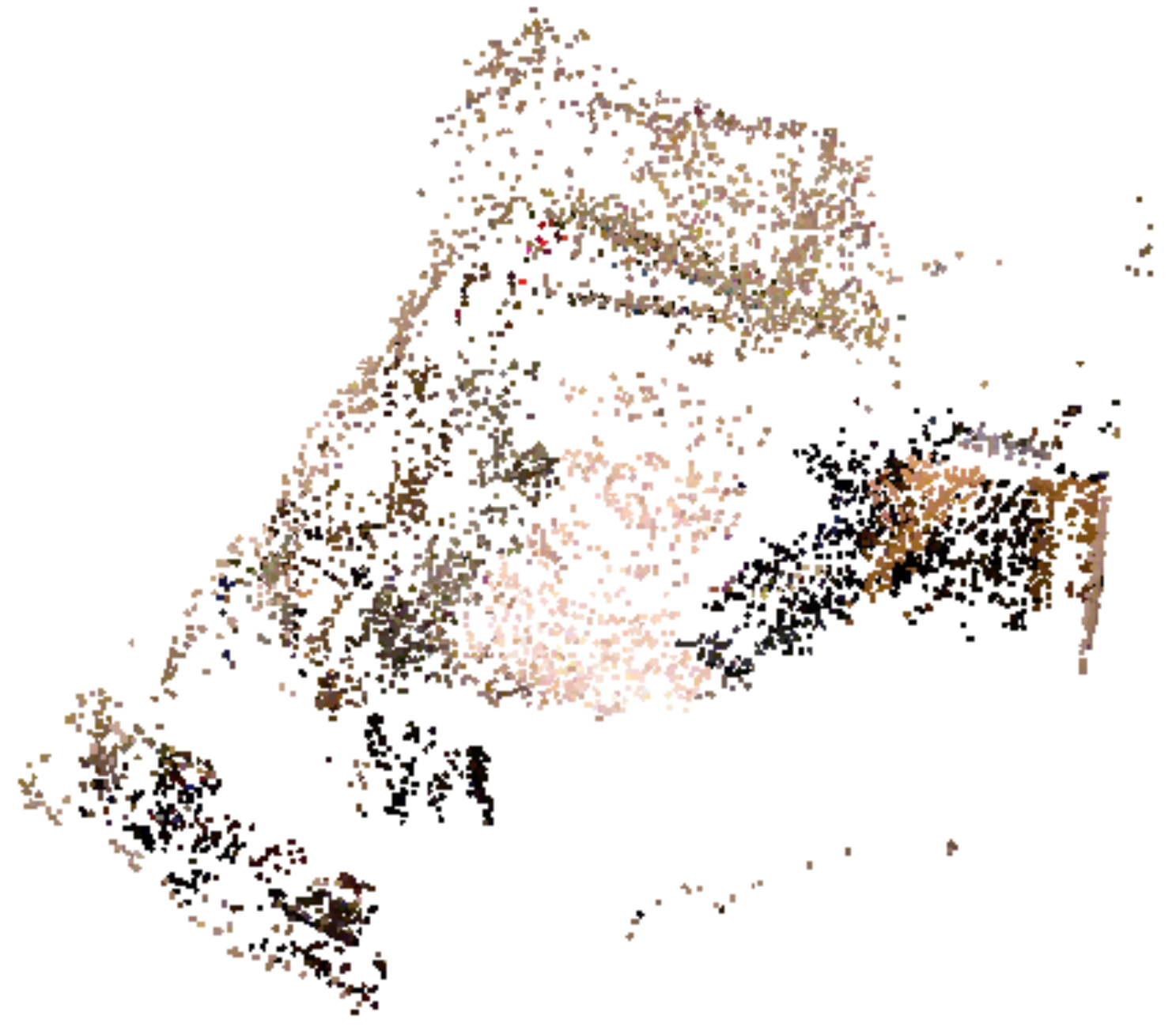} \\
  (b) sparse landmarks
  \end{minipage}
  \\
  \begin{minipage}[b]{\figWidth}\centering
  \includegraphics[width=\linewidth]{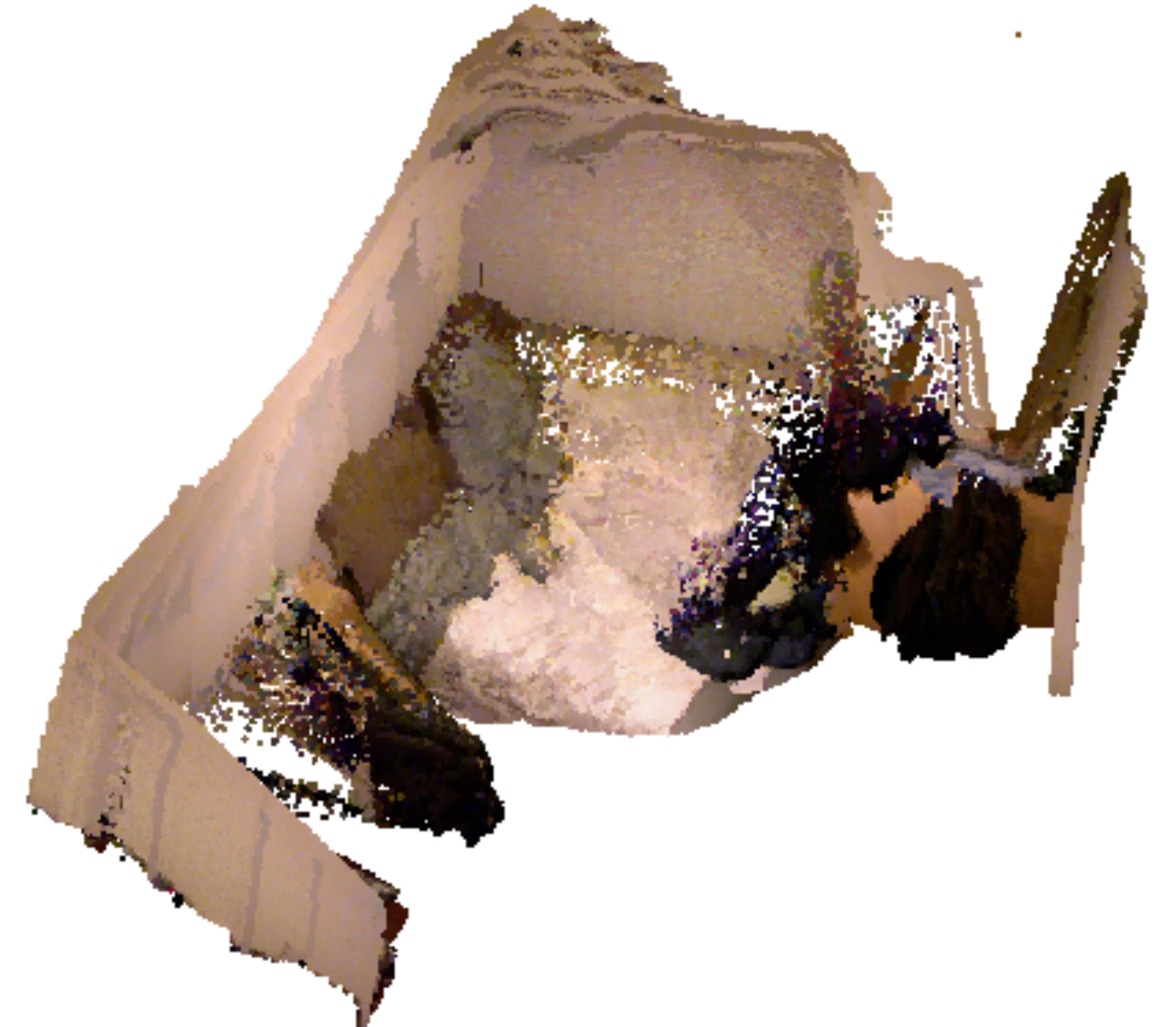} \\
  (c) ground truth map
  \end{minipage}
  &
  \begin{minipage}[b]{\figWidth}\centering
  \includegraphics[width=\linewidth]{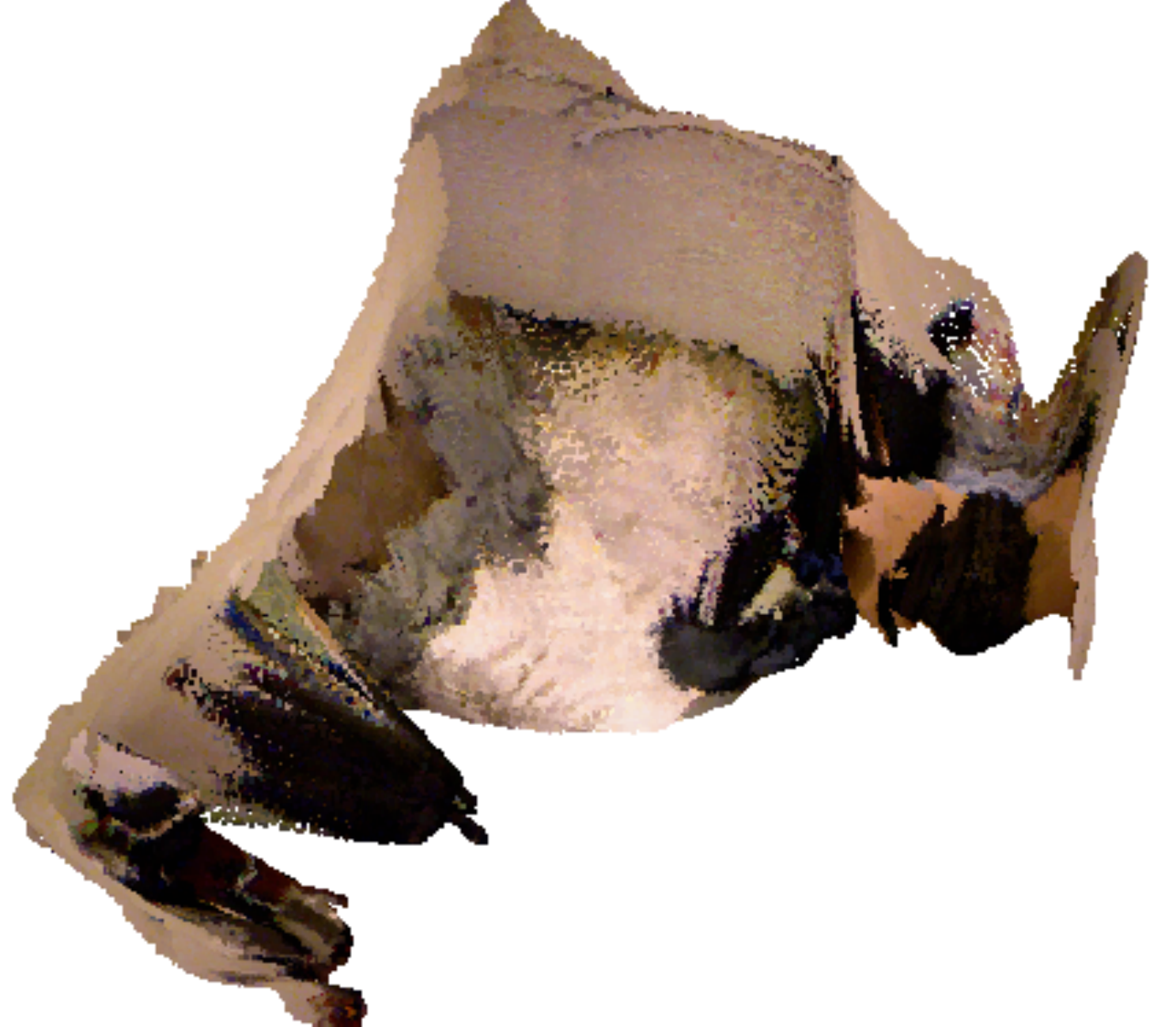} \\
  (d) prediction
  \end{minipage}
\end{tabular}
\end{minipage}
\caption{Application in sparse SLAM and visual inertial odometry (VIO) to create dense point clouds from sparse landmarks. (a) RGB (b) sparse landmarks (c) ground truth point cloud (d) prediction point cloud, created by stitching \RGBd predictions from each frame. }
\label{fig:slam}
\end{figure}

To demonstrate the effectiveness of our proposed methods, we implement a simple visual odometry (VO) algorithm with data from one of the test scenes in the \nyudepth dataset. For simplicity, the absolute scale is derived from ground truth depth image of the first frame. The 3D landmarks produced by VO are back-projected onto the RGB image space to create a sparse depth image. We use both RGB and sparse depth images as input for prediction. Only pixels within a trusted region, which we define as the convex hull on the pixel space formed by the input sparse depth samples, are preserved since they are well constrained and thus more reliable. Dense point clouds are then reconstructed from these reliable predictions, and are stitched together using the trajectory estimation from VIO. 

The results are displayed in \prettyref{fig:slam}. The prediction map resembles closely to the ground truth map, and is much denser than the sparse point cloud from VO. The major difference between our prediction and the ground truth is that the prediction map has few points on the white wall, where no feature is extracted or tracked by the VO. As a result, pixels corresponding to the white walls fall outside the trusted region and are thus removed. 

\subsection{Application: LiDAR Super-Resolution}
\label{sec:results-lidar}

\begin{figure}[htbp]
\centering
\setlength\tabcolsep{0.3mm} 
\begin{tabular}{ c }
\includegraphics[width=\linewidth, height=0.26\linewidth]{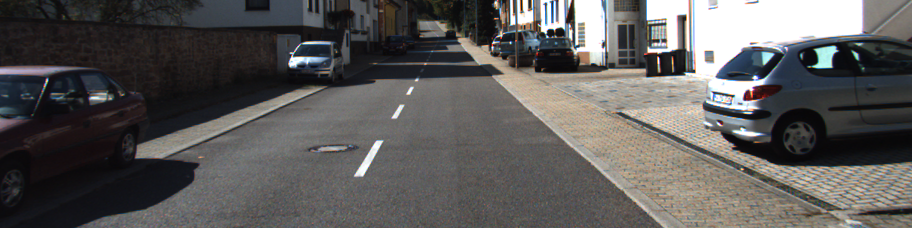} \\
\includegraphics[width=\linewidth, height=0.26\linewidth]{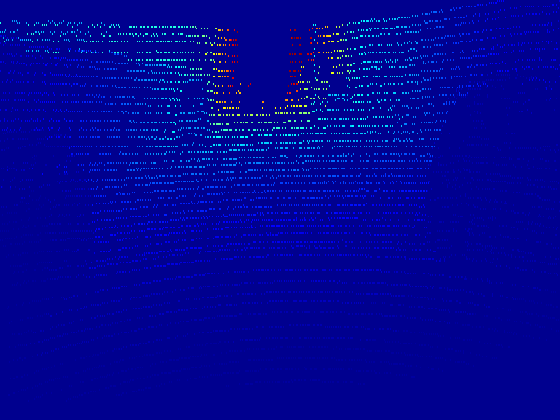} \\
\includegraphics[width=\linewidth, height=0.26\linewidth]{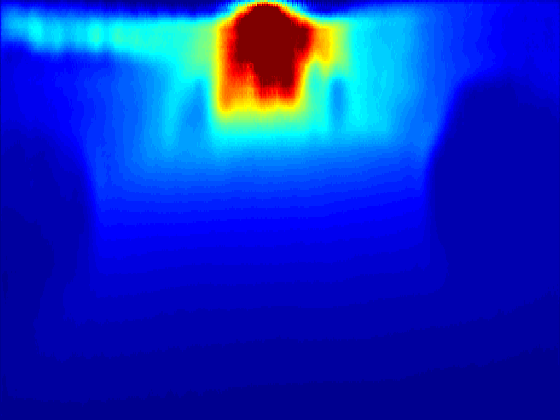} 
\end{tabular}
\caption{Application to LiDAR super-resolution, creating denser point cloud than the raw measurements. From top to bottom: RGB, raw depth, and predicted depth. Distant cars are almost invisible in the raw depth, but are easily recognizable in the predicted depth.}
\label{fig:super-resolution}
\end{figure}
We present another demonstration of our method in super-resolution of LiDAR measurements. 3D LiDARs have a low vertical angular resolution and thus generate a vertically sparse point cloud. We use all measurements in the sparse depth image and RGB images as input to our network. The average \absrel is 4.9\%, as compared to 20.8\% when using only RGB. An example is shown in \prettyref{fig:super-resolution}. Cars are much more recognizable in the prediction than in the raw scans.


\section{CONCLUSION}
We introduced a new depth prediction method for predicting dense depth images from both RGB images and sparse depth images, which is well suited for sensor fusion and sparse SLAM. 
We demonstrated that this method significantly outperforms depth prediction using only RGB images, and other existing RGB-D fusion techniques. This method can be used as a plug-in module in sparse SLAM and visual inertial odometry algorithms, as well as in super-resolution of LiDAR measurements. We believe that this new method opens up an important avenue for research into RGB-D learning and the more general 3D perception problems, which might benefit substantially from sparse depth samples. 



\section*{ACKNOWLEDGMENT}
This work was supported in part by the Office of Naval Research (ONR) through the ONR YIP program. We also gratefully acknowledge the support of NVIDIA Corporation with the donation of the DGX-1 used for this research.

\bibliographystyle{IEEEtranN}
\bibliography{references/settings,references/sparse_depth,references/rgb_prediction,references/learning,references/rgbd_cnn,references/datasets,references/slam}

\end{document}